\documentclass{article}

\usepackage[nonatbib, preprint]{neurips_2023}

\usepackage[utf8]{inputenc} 
\usepackage[T1]{fontenc}    
\usepackage{hyperref}       
\usepackage{url}            
\usepackage{booktabs}       
\usepackage{amsfonts}       
\usepackage{nicefrac}       
\usepackage{microtype}      
\usepackage{xcolor}         
\usepackage{amsmath}
\usepackage{graphicx}
\usepackage{multirow}
\usepackage{wrapfig}
\usepackage{mathrsfs}
\usepackage{tcolorbox}
\usepackage{url}
\usepackage{utfsym}
\usepackage{cleveref}

\crefname{figure}{Figure}{Figures}
\Crefname{figure}{Figure}{Figures}

\title{Chat-3D: Data-efficiently Tuning Large Language Model for Universal Dialogue of 3D Scenes}
\author{Zehan Wang\thanks{Equal contribution} \hspace{1em} Haifeng Huang\footnotemark[1] \hspace{1em} Yang Zhao \hspace{1em} Ziang Zhang  \hspace{1em} Zhou Zhao\thanks{Corresponding author}\\
Zhejiang University\\
{\tt\small \{wangzehan01, huanghaifeng, zhaozhou\}@zju.edu.cn}
}

\begin{document}

\maketitle


\begin{abstract}
3D scene understanding has gained significant attention due to its wide range of applications. However, existing methods for 3D scene understanding are limited to specific downstream tasks, which hinders their practicality in real-world applications. This paper presents Chat-3D, which combines the 3D visual perceptual ability of pre-trained 3D representations and the impressive reasoning and conversation capabilities of advanced LLMs to achieve the first universal dialogue systems for 3D scenes. Specifically, we align 3D representations into the feature space of LLMs, thus enabling LLMs to perceive the 3D world. Given the scarcity of 3D scene-text data, we propose a three-stage training strategy to efficiently utilize the available data for better alignment. To enhance the reasoning ability and develop a user-friendly interaction scheme, we further construct a high-quality object-centric 3D instruction dataset and design an associated object-centric prompt. Our experiments show that Chat-3D achieves an impressive ability to comprehend diverse instructions for 3D scenes, engage in intricate spatial reasoning, and incorporate external knowledge into its responses. Chat-3D achieves a 75.6\% relative score compared with GPT-4 on the constructed instruction dataset.
The project page is \url{https://chat-3d.github.io/}
\end{abstract}

\section{Introduction}
3D vision is an important way for robots to perceive the rich semantic and spatial information of the real world. 3D scene understanding~\cite{azuma2022scanqa, ma2022sqa3d, chen2020scanrefer, achlioptas2020referit3d, chen2021scan2cap} has garnered increasing attention in recent years, owing to its broad range of applications in human-robot interaction, metaverse, robotics, and embodied intelligence. However, current methods~\cite{wang20233drp, wang2023distilling, yang2021sat, jiao2022more, yuan2022x, parelli2023clip} are limited in addressing specific downstream tasks, such as captioning and question answering, while lacking the ability to engage in general dialogue regarding a 3D scene, restricting their practicality in various real-world tasks. A universal dialogue system for 3D scenes is an imperative component of high-level intelligent robots.

The general dialogue system for 3D scenes requires two kinds of abilities: 3D perception and reasoning. Recently, several studies~\cite{yu2022point, pang2022masked, wang2021unsupervised, zhang2022point, xue2023ulip, liu2023openshape} on pre-trained 3D representations shows impressive performance in 3D perception. However, the reasoning ability for the 3D world remains constrained owing to the scarcity of reasoning and describing data for 3D scenes.

Large language models (LLMs)~\cite{chiang2023vicuna, openai2023gpt4, touvron2023llama, chowdhery2022palm}, on the other hand, exhibit remarkable prowess in complex reasoning and open-domain conversations. Moreover, recent methods~\cite{li2023videochat, liu2023visual, zhao2023bubogpt, zhang2023video, zhu2023minigpt} attempt to extend LLMs to image and video fields. These works typically adopt a two-stage training scheme: Firstly, the visual representations are aligned into the word embedding space of LLMs by leveraging large-scale image-text and video-text datasets~\cite{lin2014microsoft, sharma2018conceptual, changpinyo2021conceptual, schuhmann2021laion, schuhmann2022laion, bain2021frozen, miech2019howto100m, xu2016msr}. Secondly, they enhance the reasoning capabilities of LLMs regarding visual concepts by fine-tuning on the instruction datasets.

Despite the success of image and video understanding fields, introducing LLMs to perceive 3D scenes faces two challenges: 1) Compared to the millions or even billions of image-text and video-text data~\cite{sharma2018conceptual, changpinyo2021conceptual, schuhmann2021laion, schuhmann2022laion, bain2021frozen}, the 3D scene-text data~\cite{achlioptas2020referit3d, chen2020scanrefer} is limited. Consequently, in the low-resource scenarios, the commonly used two-stage training scheme in previous multi-modal LLMs is less effective in aligning pre-trained 3D representations to the feature space of LLMs. 2) 3D scenes always encompass a greater number of objects compared to an image or a video clip. Thus, the common questions or instructions in images and videos are more susceptible to ambiguity in 3D scenes. Consider a simple question like "What is in front of this chair?" on a 3D scene that contains multiple chairs. The dialogue model cannot understand which specific chair the user is asking about, and uniquely describing an object (the chair) in question is often difficult and user-unfriendly due to the complex object relations.

In this paper, we propose Chat-3D, the first attempt to extend the reasoning and conversation capabilities of LLMs to 3D scene understanding. We employ a three-stage training scheme to more efficiently utilize the limited data. Specifically, in the first stage, we directly align the features of 3D objects with the word embeddings of their class names. In the second stage, we learn a 3D object relation module via 3D scene-text data to capture semantic information about the whole 3D scene. Finally, in the third stage, we further tune the model with a high-quality instruction dataset. To further enhance the reasoning ability of Chat-3D, we construct the instruction dataset via an object-centric scheme, which means all instructions are related to a specific object. Combining our object-centric prompt, users can effortlessly select the object in the scene they want to engage in a dialogue about, without the need to uniquely describe the specific object in their instructions.

In summary, our contributions can be summarized as follows:

(1) We build the first universal dialogue system for 3D scenes, leveraging the advanced visual perception capabilities of 3D pre-trained models, in conjunction with the powerful reasoning and open-domain conversational abilities of LLMs.

(2) We introduce a new three-stage training scheme for multi-modal LLM, enabling the model to progressively transition from learning individual object attributes to capturing complex spatial object relations. This approach effectively improves the quality of dialogue with limited available data.

(3) We construct a high-quality object-centric 3D instruction dataset including diverse dialogues about object attributes, positions, relationships, functionalities, placement suggestions, and detailed descriptions within 3D scenes. We propose a corresponding object-centric prompt approach to provide a user-friendly interaction method.

(4) Our experiments demonstrate that Chat-3D exhibits remarkable capabilities in universal dialogue and spatial reasoning based on 3D scenes. We also employ quantitative comparison to evaluate the effectiveness of our three-stage training scheme and instruction dataset.

\section{Related Work}
\paragraph{3D Representation Learning.} 3D point cloud is a fundamental visual modality. Recently, numerous attempts are made to learn discriminative and robust representations for point cloud objects. PointBERT~\cite{yu2022point}, Point-MAE~\cite{pang2022masked}, Transformer-OcCo~\cite{wang2021unsupervised}, and point-m2ae~\cite{zhang2022point} employ self-supervised learning approaches to extract meaningful representations of 3D objects from unlabeled point cloud data. Another series of works aims to extend representation from other modalities to 3D. For instance, ULIP~\cite{xue2023ulip} and openshape~\cite{liu2023openshape} construct (3D-image-text) triplets to align point clouds within the CLIP~\cite{radford2021learning, cherti2023reproducible} representation space, while I2P-MAE~\cite{zhang2023learning} and ACT~\cite{dong2023act} learn 3D representations from image pre-trained models~\cite{dosovitskiy2020image, he2016deep}. These powerful 3D representations can effectively capture the detailed information of a 3D object. In Chat-3D, we segment the 3D scene into objects and extract features for each object, which yields a set of object features to represent the 3D scene and serves as a prerequisite for an object-centric interactive approach.

\paragraph{3D-Language Tasks.} The interaction between 3D point clouds and natural language has wild applications and has garnered increasing attention recently. 
3D captioning~\cite{chen2021scan2cap, chen2020scanrefer, achlioptas2020referit3d} focuses on generating descriptions of a specific object in a 3D scene. In 3D visual question answering~\cite{azuma2022scanqa}, the model is required to answer questions based on the visual content of the 3D scene, while the more complex 3D situated question answering~\cite{ma2022sqa3d} requires the model to understand agent's situation (position, orientation, etc.) in a 3D scene as described by text, reason about the surrounding environment. Different from vision-language tasks~\cite{kazemzadeh2014referitgame,  krishna2017dense, goyal2017making, antol2015vqa, lin2014microsoft, grauman2022ego4d} and methods~\cite{li2022blip, li2023blip, li2021align, lin2022swinbert, deng2021transvg, wang2023scene} based on images and videos, these 3D-language tasks and corresponding methods place more emphasis on spatial reasoning and the possible interaction between agents and scenes. Despite the significant progress made in this field, existing methods still focus on improving isolated task-specific models, without exploring a unified dialogue system.

\paragraph{Multi-modal Large Language Models.} Recently, Large Language Models showcase remarkable abilities in complex reasoning and conversational communication with humans. To extend the knowledge, reasoning, and conversation abilities acquired from vast amounts of text data to more modalities, some studies~\cite{li2023videochat, liu2023visual, zhao2023bubogpt, zhang2023video, zhu2023minigpt} attempt to instruction tune LLMs for multimodal learning. Specifically, these works first use the caption learning objective to learn the aligning of visual features with pre-trained LLMs from large-scale vision-language paired data. Then, a high-quality instruction dataset is utilized to further enhance the LLMs' comprehension of the visual world. However, in the 3D-Language field, 3D scene-text pairs are scarce. Thus the indirect aligning method is unreliable and incomplete for 3D representations and pre-trained LLMs. To mitigate this issue, we propose a more data-efficient three-stage tuning scheme that establishes a more direct learning stage for alignment, reduces the annotation requirements, and provides a smooth learning curve.

\section{Methods}

\subsection{Architecture}
\label{sec:arch}
\begin{wrapfigure}{r}{6.5cm} 
  \centering
  \includegraphics[width=6.5cm]{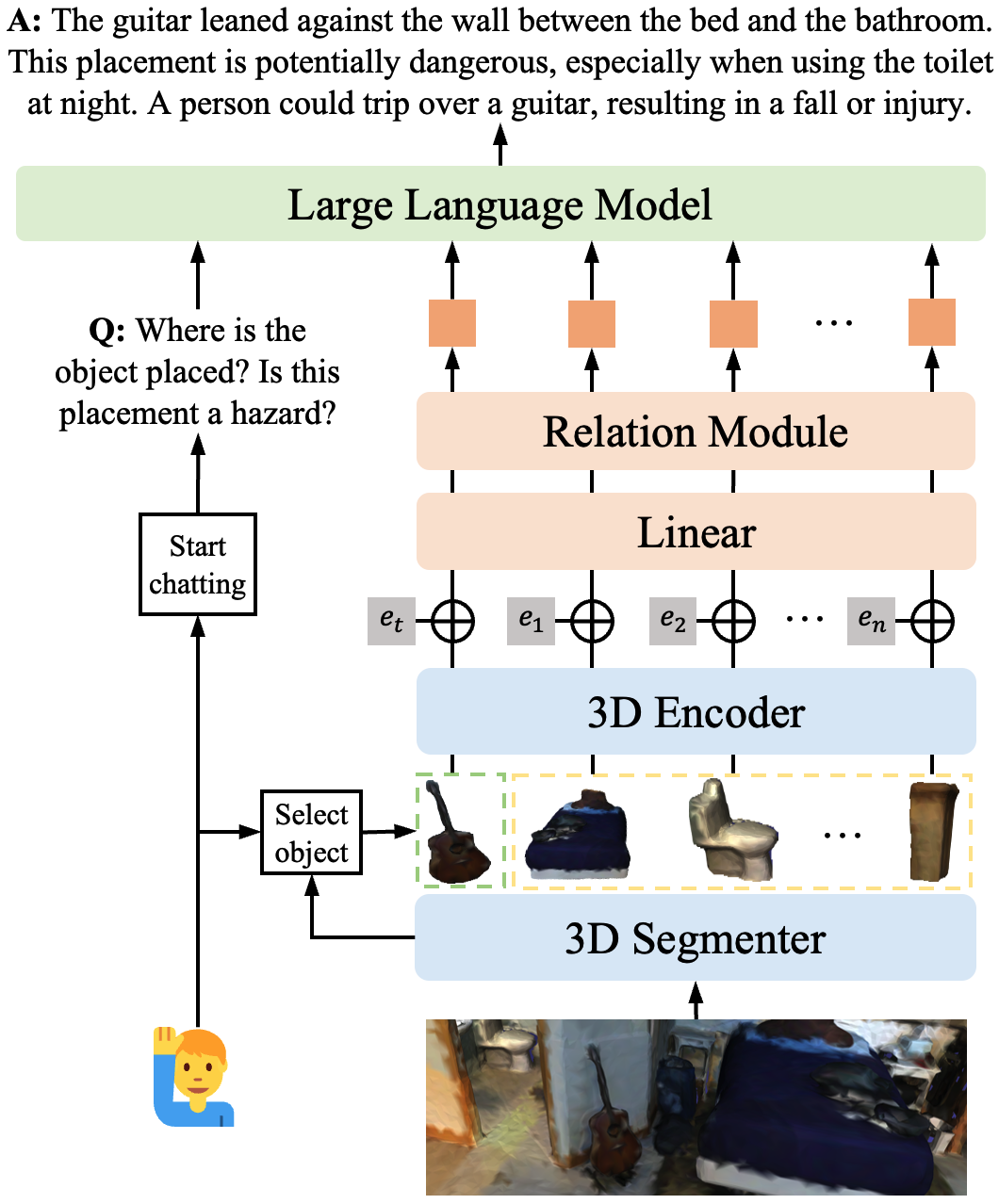}
  \caption{The overall architecture of Chat-3D.}
  \label{fig:framework}
\end{wrapfigure}

Chat-3D aims to create a universal dialogue system for 3D scenes by aligning 3D representations with pre-trained LLM~\cite{touvron2023llama}. The overall network architecture is illustrated in Figure \ref{fig:framework}. 

For the input 3D scene $S$, we first use a 3D object segmentation model~\cite{jiang2020pointgroup, misra2021end, qi2019deep} or ground truth annotations to segment it into objects. Then, users can select the specific object they want to engage in dialogue. The selected target object is denoted as $o_t$ and other objects in the scene are represented as $O_s = [o_1, o_2, \dots, o_{n_s}]$, where $n_s$ is the number of other objects in the 3D scene. For each object, we use a pre-trained 3D point encoder $g(\cdot)$ to extract features, Besides, we further incorporate extra object attributes (e.g. color, size, location) into these object features by a projector $f_e(\cdot)$ to enrich semantic information. These 3D object features are projected to the word embedding space of pre-trained LLM via a projector $f_a(\cdot)$. The process of 3D object feature extraction and mapping can be expressed as:
\begin{equation}
    \label{eq:object}
    \mathbf{z}_i = f_a(g(o_i) + \mathbf{e}_i), \ \operatorname{with} \mathbf{e}_i = f_e([\mathbf{c}_i; \mathbf{s}_i; \mathbf{l}_i])
\end{equation}
where $i \in [t, 1, 2, \dots, n_s]$, and $\mathbf{c}_i, \mathbf{s}_i, \mathbf{l}_i \in \mathbb{R}^3$ respectively represent the RGB value, bounding box size, and location for the $i$-th object. The extracted 3D features of target object and other objects are denoted as $\mathbf{z}_t$ and $\mathbf{Z}_s = [\mathbf{z}_1, \mathbf{z}_2, \dots, \mathbf{z}_{n_s}]$.

Furthermore, we further introduce a relation module $t(\cdot)$ for capturing complex relations between objects. The features of objects then interact with each other to provide additional object relation information about the scene.
\begin{equation}
\label{eq:scene}
[\mathbf{\hat{z}}_t, \mathbf{\hat{z}}_1, \mathbf{\hat{z}}_2, \dots, \mathbf{\hat{z}}_{n_s}] = r([\mathbf{z}_t, \mathbf{z}_1, \mathbf{z}_2, \dots, \mathbf{z}_{n_s}])   
\end{equation}

The representations of a 3D scene are provided as $\mathbf{\hat{z}}_t \in \mathbb{R}^d$, $[\mathbf{\hat{z}}_1, \mathbf{\hat{z}}_2, \dots, \mathbf{\hat{z}}_{n_s}] \in \mathbb{R}^{n_s \times d}$, and $d$ is the dimension of hidden states in the pre-trained LLMs.

Lastly, to facilitate user-friendly interaction between our system and users, we design an object-centric prompt as: \textit{\#\#\#Human: <target> $\mathbf{\hat{z}}_t$ </target> <scene> $\mathbf{\hat{z}}_1, \mathbf{\hat{z}}_2, \dots, \mathbf{\hat{z}}_{n_s}$ </scene> <instruction> \#\#\#Assistant:}. Through this prompt, the LLM can comprehend the specific object the user wants to discuss and generate responses based on the 3D visual information and the given instructions.

\subsection{Three-stage Training}
Previous multi-modal LLMs~\cite{li2023videochat, liu2023visual, zhao2023bubogpt, zhang2023video, zhu2023minigpt} primarily follow a two-stage training scheme. In the first stage, LLMs take inputs from visual modality and learn to generate corresponding captions. The large-scale image- and video-text datasets allow comprehensive alignment between visual representations and the word embedding space of LLM. In the second stage, the model is fine-tuned with a high-quality instruction dataset, thereby further enhancing the perceptual and reasoning abilities.

However, in the 3D understanding field, the 3D scene-text data is significantly less than image- or video-text data. For example, the commonly used ScanRefer~\cite{chen2020scanrefer} dataset, which provides descriptions for ScanNet~\cite{dai2017scannet}, only contains 36,655 captions for training. In contrast, the datasets used for the first stage training in previous multi-modal LLM methods are million-level or even billion-level, such as CC3M~\cite{sharma2018conceptual}, CC12M~\cite{changpinyo2021conceptual}, LAION-400M~\cite{schuhmann2021laion}, LAION-5B~\cite{schuhmann2022laion} and WebVid-10M~\cite{bain2021frozen}. Considering the scarcity of 3D scene-text data, we propose a more data-efficient three-stage training approach, which divides the process of aligning 3D features with the pre-trained LLM into two progressive stages: 3D object alignment and 3D scene alignment.

\paragraph{Stage 1: 3D Object Alignment} The first stage is designed to learn the alignment between the representation of individual 3D objects and pre-trained LLM. Given a 3D object and its annotated category, the 3D object is encoded into a representation $\mathbf{z} \in \mathbb{R}^d$ according to Equation \ref{eq:object}. Its category name is encoded into a word embedding $\mathbf{y} \in \mathbb{R}^d$ using the tokenizer of the pre-trained LLM. By maximizing the cosine similarity between the corresponding $\mathbf{z}$ and $\mathbf{y}$, we can learn projectors $f_e(\cdot)$ and $f_a(\cdot)$ that effectively inject the 3D object representations into the word embedding space of LLM.

The advantage of Stage 1 is three-fold: 1) Compared to learning alignment through captioning objective, maximizing the similarity between representations provides a more direct learning objective for alignment, which can achieve more efficient alignment in low-resource scenarios. 2) Stage 1 enables the utilization of 3D point cloud object classification datasets, such as ShapeNet~\cite{chang2015shapenet}, ScanObjectNN~\cite{uy2019revisiting}, and Objaverse~\cite{deitke2023objaverse}, which enhances the model's generalization performance on diverse real-world objects. 3) The introduction of Stage 1 offers a smoother learning curve for comprehending complex 3D scenes. The model progressively transitions from learning individual object attributes to capturing intricate spatial object relations.

\paragraph{Stage 2: 3D Scene Alignment} After aligning individual 3D object feature with pre-trained LLM, Stage 2 takes a step further by integrating the entire 3D scene into LLM. The training data is sourced from the ScanRefer dataset, which provides annotations for objects in a scene primarily based on their spatial relationships. Considering a 3D scene, which can be segmented into object set $[o_1, o_2, \dots, o_n]$, we sequentially select each object as target objects and construct the input for LLM according to the methodology discussed in Section \ref{sec:arch}. The instruction in prompts requests the model to generate a brief description of the target object within the 3D scene. The learning objective is to generate a description that aligns with the description provided by the ScanRefer dataset for the target object, and only the two projectors $f_e(\cdot)$, $f_a(\cdot)$ and the relation module $r(\cdot)$ are learnable in this stage.

\paragraph{Stage 3: Instruction Tuning} For enhancing the reasoning ability about 3D world, we curate a high-quality instruction dataset which comprises rich and detailed instructions. By tuning Chat-3D on this dataset, we further enhance its capability to comprehend diverse instructions, generate imaginative and contextually appropriate responses, engage in intricate spatial reasoning, and effectively incorporate external knowledge into its responses.

\section{Object-centric Instruction Dataset}
\label{sec:instruction}
\begin{table}[h]
\begin{tcolorbox}[colback=gray!10,
                  colframe=black,
                  width=\textwidth,
                  arc=1mm, auto outer arc,
                  boxrule=0.5pt,
                 ]
\textcolor{cyan}{\textbf{Caption of the target object:}}

Descriptions: ["There is a single white armchair. placed next to the window of the room.", "The sofa chair is the corner chair. lying parallel to the wall. a small table with the lamp is present beside the chair.", "This is a white sofa chair. it is under a window.", "This is a white armchair. is next to a lamp.", "This is the corner sofa chair. a small table with a lamp can be seen near this chair."]

\textcolor{cyan}{\textbf{Categories and locations of target object and its 10 neighbors:}}

Described object: \{sofa chair:[-1.31, 3.15, 0.59]\}; Neighbor objects: \{window:[-1.12, 4.12, 1.59], table:[0.86, 1.61, 0.38], doorframe:[-2.25, 0.67, 1.27], windowsill:[0.88, 3.97, 0.98], windowsill:[-1.32, 3.93, 0.91], sofa chair:[0.98, 3.35, 0.71], window:[1.16, 4.18, 1.73], pillow:[1.35, 0.29, 0.46], table:[-0.15, -2.66, 0.26], tv:[-2.2, -0.55, 1.52]\}
\end{tcolorbox}
\caption{An example of textualizing an object in a 3D scene}
\label{tab:example}
\end{table}

\begin{table}[h]
\begin{tcolorbox}[colback=gray!10,
                  colframe=black,
                  width=\textwidth,
                  arc=1mm, auto outer arc,
                  boxrule=0.5pt,
                 ]
You are an AI 3D visual assistant, and you are seeing an object in a 3D scene. What you see is provided with several sentences, describing the same object you are looking at, and the position of surrounding objects in the 3D scene to represent the content of the 3D scene. Based on these descriptions of this object and the location of surrounding objects in the 3D scene, summary and describe the placement, function of this object, and how a person can access this object in detail as if you are in the 3D scene. 

Importantly, do not mention any specific spatial coordinate values. The description should be more than 150 words and less than 200 words.
\end{tcolorbox}
\caption{Prompt for descriptive object-centric captions.}
\label{tab:prompt1}
\end{table}

The complex object relationships and intricate interactions between agents and scenes impose elevated demands on reasoning capabilities. To enhance the reasoning ability pertaining to 3D world, we construct a high-quality object-centric instruction dataset based on the annotations in ScanRefer. Specifically, we leverage the remarkable reasoning and summarizing capabilities of ChatGPT to automatically generate descriptive and detailed captions as well as diverse conversations centered around specific objects within 3D scenes.

\paragraph{Object-centric Descriptive Captions.} 
ScanRefer annotates multiple captions for objects in a 3D scene based on their attributes and spatial relationships. We employ ChatGPT to summarize and rewrite these short captions into imaginative paragraphs. To facilitate ChatGPT's comprehension of the 3D scene, we also textualize the 3D scene as shown in Table \ref{tab:example}, providing the categories and XYZ coordinates of the target object and its ten nearest objects in the scene. Furthermore, we propose a prompt to request ChatGPT to focus on perceiving and reasoning about the object relations and agent interactions as exemplified in Table \ref{tab:prompt1}. During dataset construction, we initially manually annotated several examples and randomly selected two of them as in-context examples to guide the generation of ChatGPT. One example of the generated descriptive object-centric caption is shown in Table \ref{tab:describe}.

\begin{table}[h]
\begin{tcolorbox}[colback=gray!10,
                  colframe=black,
                  width=\textwidth,
                  arc=1mm, auto outer arc,
                  boxrule=0.5pt,
                 ]

\begin{center}
\includegraphics[width=0.6\textwidth]{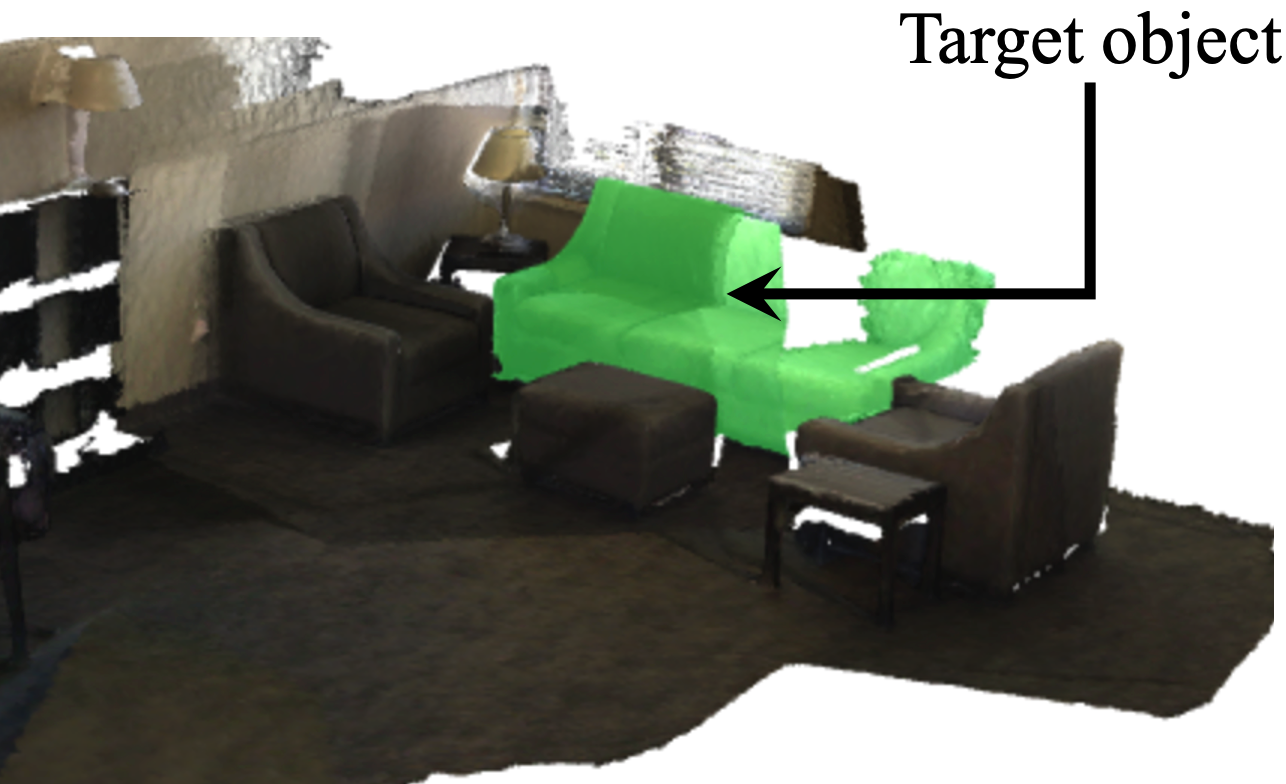}
\end{center}

\textcolor{cyan}{\textbf{Detailed description:}} The object is a brown couch with three seats, positioned on the left corner of the room. It is next to a gold lamp, which serves as a source of lighting and complements the overall aesthetics of the area. The couch is accompanied by an ottoman placed in front of it, providing a comfortable footrest and enhancing the seating experience.

The presence of two armchairs nearby indicates that this space is designed for relaxation and socializing. The grey-colored couch with three cushions exudes a modern and stylish vibe, adding to the room's visual appeal. The strategic placement of the couch and ottoman encourages a cozy and inviting atmosphere, ideal for leisurely activities and gatherings.

This setup creates a comfortable seating arrangement, promoting interaction and conversation among individuals using the couch and armchairs. The nearby window adds natural light to the area, enhancing the ambiance. The presence of an end table with a lamp contributes to the functionality of the space by providing a convenient spot for placing items and lighting.

In summary, the brown couch with its ottoman and neighboring armchairs forms a functional and aesthetically pleasing seating arrangement. The combination of the couch's design, the presence of the ottoman, and the proximity to the end table with a lamp ensures a comfortable and enjoyable experience for individuals using this part of the room.

\end{tcolorbox}
\caption{Example of descriptive object-centric caption.}
\label{tab:describe}
\end{table}

\paragraph{Object-centric Conversations.} To enhance the capability of handling diverse instructions and general conversations, we further require ChatGPT to autonomously generate multi-turn dialogues in a self-questioning and self-answering format based on the brief captions of the target object and the textualized 3D scene.

\section{Experiments}

\subsection{Implementation Details}
During the training phase, we directly use ground truth annotations (point cloud and extra attributes) of each object in the 3D scene for better training quality. We employ the pre-trained Point-Bind\cite{pointbind} model with Point-BERT\cite{yu2022point} architecture as $g(\cdot)$ to extract features for each object. Meanwhile, we use a linear layer as $f_e(\cdot)$ to incorporate extra attributes (such as color, size, and location) into the extracted features. Then, a two-layer MLP serves as $f_a(\cdot)$ to map these 3D object features to the word embedding space of the pre-trained LLM, and the relation module $r(\cdot)$ is implemented using a one-layer vanilla transformer encoder. It is worth mentioning that the relation module is zero-initialized, thereby preserving the information learned in Stage 1 when Stage 2 begins. The chosen LLM for our experiment is a Vicuna 7B model\cite{chiang2023vicuna}, which is fine-tuned from the LLaMA base model\cite{touvron2023llama}.

\begin{table}[b]
\begin{tabular}{ccc|ccc}
\toprule
\multirow{2}{*}{\begin{tabular}[c]{@{}c@{}}Training\\ scheme\end{tabular}} &
  \multicolumn{2}{c|}{Training Data} &
  \multicolumn{2}{c}{Evaluate Set} &
  \multirow{2}{*}{Overall} \\
            & Conversation & Detailed Caption & Conversation & Detailed Caption &      \\ \midrule
Three-Stage & \checkmark            & \checkmark                   & 84.0         & \textbf{67.6}                & \textbf{75.6} \\
Two-Stage   & \checkmark            & \checkmark                   & 78.0         & 56.2                & 67.0 \\
Three-Stage & \checkmark            & \scalebox{0.75}{\usym{2613}}                   & \textbf{84.7}         & 50.1                & 67.3 \\
Three-Stage & \scalebox{0.75}{\usym{2613}}            & \checkmark                    & 81.5         & 62.7                & 71.9 \\
Three-Stage & \scalebox{0.75}{\usym{2613}}            & \scalebox{0.75}{\usym{2613}}                   & 53.4         & 41.6                & 47.4 \\ \bottomrule
\end{tabular}
\caption{Relative scores for different settings.}
\label{tab:score}
\end{table}

\subsection{Quantitative Analysis}
In order to quantitatively evaluate the universal dialogue ability of Chat-3D and analyze the effect of the three-stage training scheme and our instruction dataset, we adopt GPT-4~\cite{openai2023gpt4} to measure the quality of our Chat-3D's generated responses following LLaVA~\cite{liu2023visual} and miniGPT4~\cite{zhu2023minigpt}. Specifically, we randomly select 30 scenes from the ScanRefer validation set and randomly choose one object as the target object for each scene. We employ the instruction dataset construction methodology described in Section~\ref{sec:instruction} and Chat-3D respectively to generate responses under the same scene and instruction inputs. After that, we input the textualized scene, instructions, and the two kinds of generated responses into GPT-4 and request GPT-4 to provide an overall score on a scale of 1 to 10 for each response based on its helpfulness, relevance, accuracy, and level of detail. A higher score indicates a higher quality of response. 

In Table \ref{tab:score}, we study the effectiveness of the instruction dataset and compare the Chat-3D trained via our three-stage training scheme and the two-stage training method used in previous methods~\cite{li2023videochat, liu2023visual, zhao2023bubogpt, zhang2023video, zhu2023minigpt}. First, our three-stage training scheme significantly outperforms the previous two-stage method by 8.6 points, demonstrating the data efficiency of our three-stage training method in the low-resource setting. Second, by comparing different combination settings of the instruction dataset, we observe that incorporating conversation data leads to a higher improvement in conversation tests, while integrating detailed caption data enhances performance in detailed caption tests. By utilizing all the data together, our model demonstrates proficiency in both conversation and detailed caption tasks, ultimately achieving the highest overall score.

\subsection{Qualitative Comparisons \& Analysis}
In section, we provide visualization examples of conversations about 3D scenes with Chat-3D. From these cases, we mainly study the perception, reasoning, and dialogue capabilities of Chat-3D. Besides, we further compare Chat-3D with 2D multimodal LLM methods such as MiniGPT-4~\cite{zhu2023minigpt}, LLaVA~\cite{liu2023visual}, and mPLUG-owl~\cite{ye2023mplug} to demonstrate the advantages and necessity of developing a specific multi-modal LLM for 3D scenes.

\begin{figure}[tb]
  \centering
  \includegraphics[width=\textwidth]{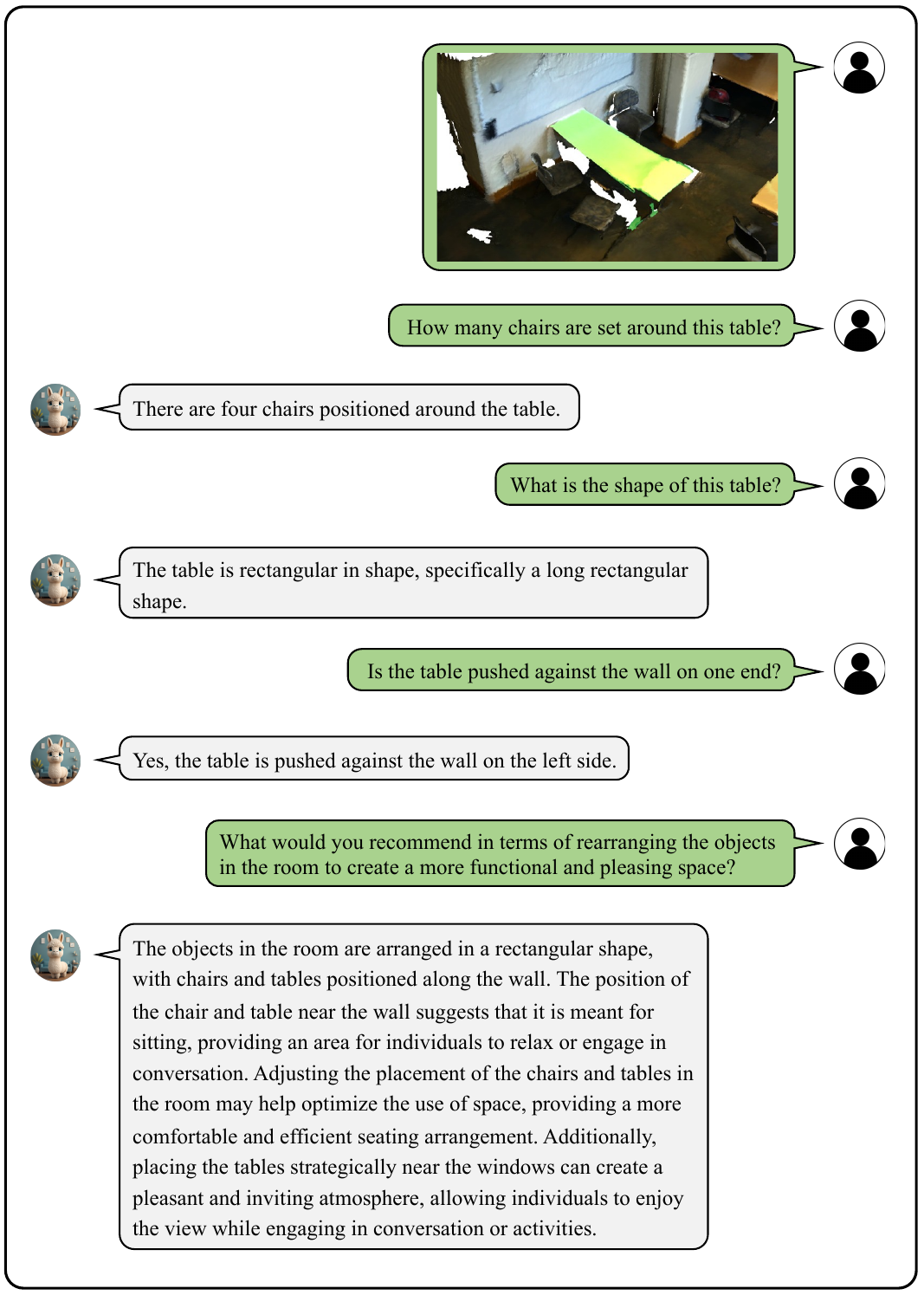}
  \caption{Example 1 of Chat-3D conversation.}
  \label{fig:chat_example1}
\end{figure}

\begin{figure}[tb]
  \centering
  \includegraphics[width=\textwidth]{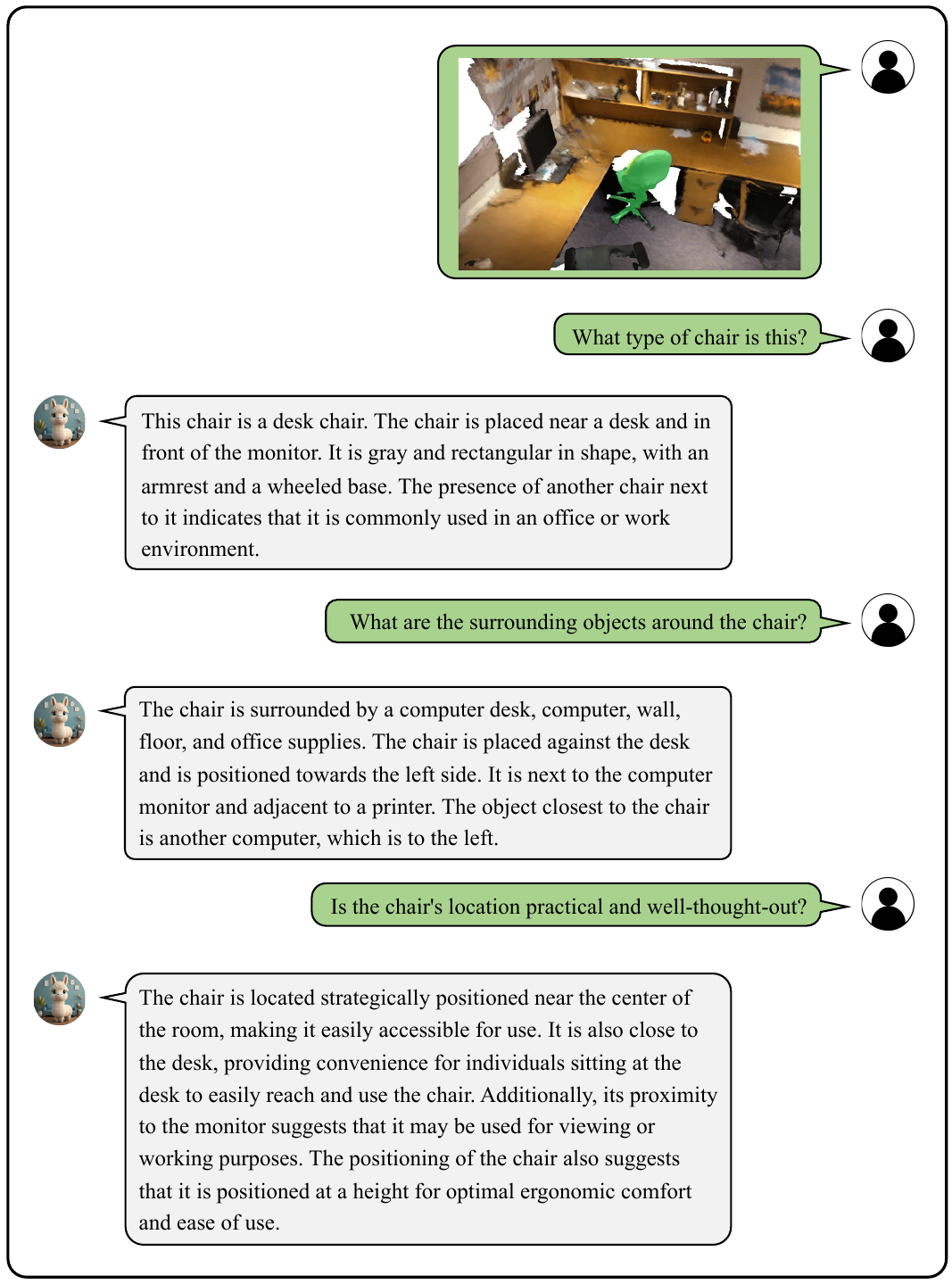}
  \caption{Example 2 of Chat-3D conversation.}
  \label{fig:chat_example2}
\end{figure}

\begin{figure}[tb]
  \centering
  \includegraphics[width=0.9\textwidth]{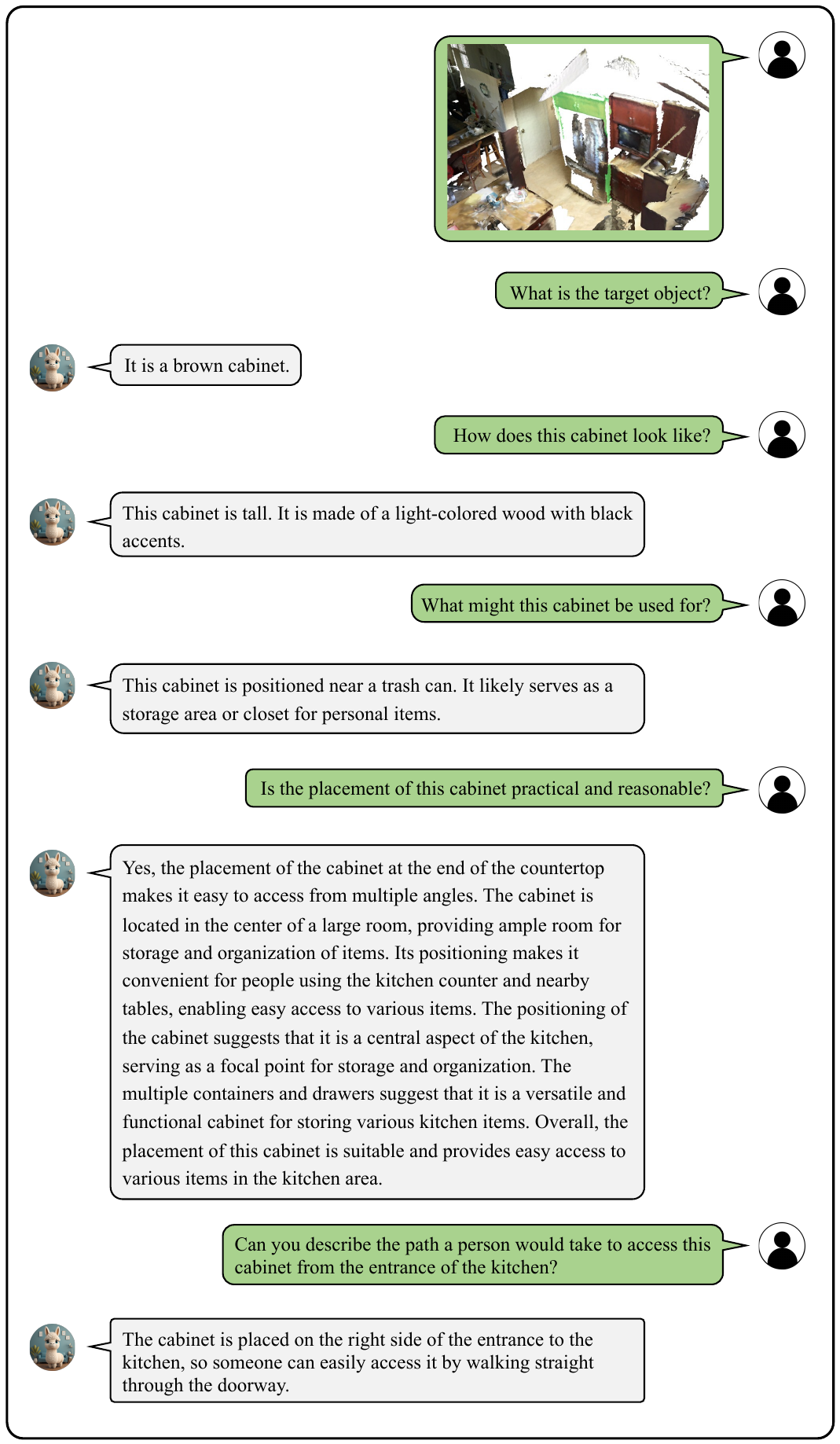}
  \caption{Example 3 of Chat-3D conversation.}
  \label{fig:chat_example3}
\end{figure}

\begin{figure}[tb]
  \centering
  \includegraphics[width=\textwidth]{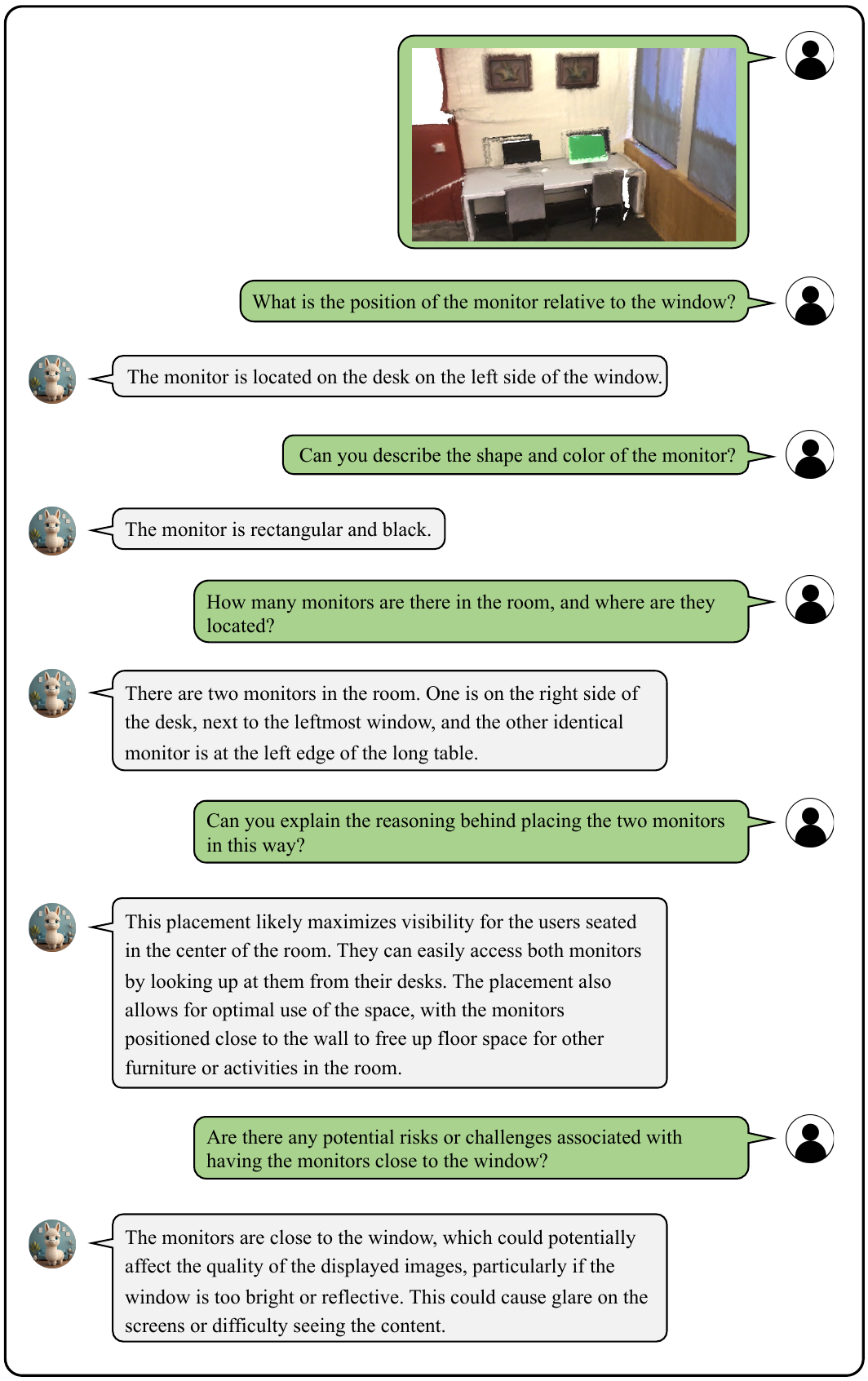}
  \caption{Example 4 of Chat-3D conversation.}
  \label{fig:chat_example4}
\end{figure}

\begin{figure}[tb]
  \centering
  \includegraphics[width=\textwidth]{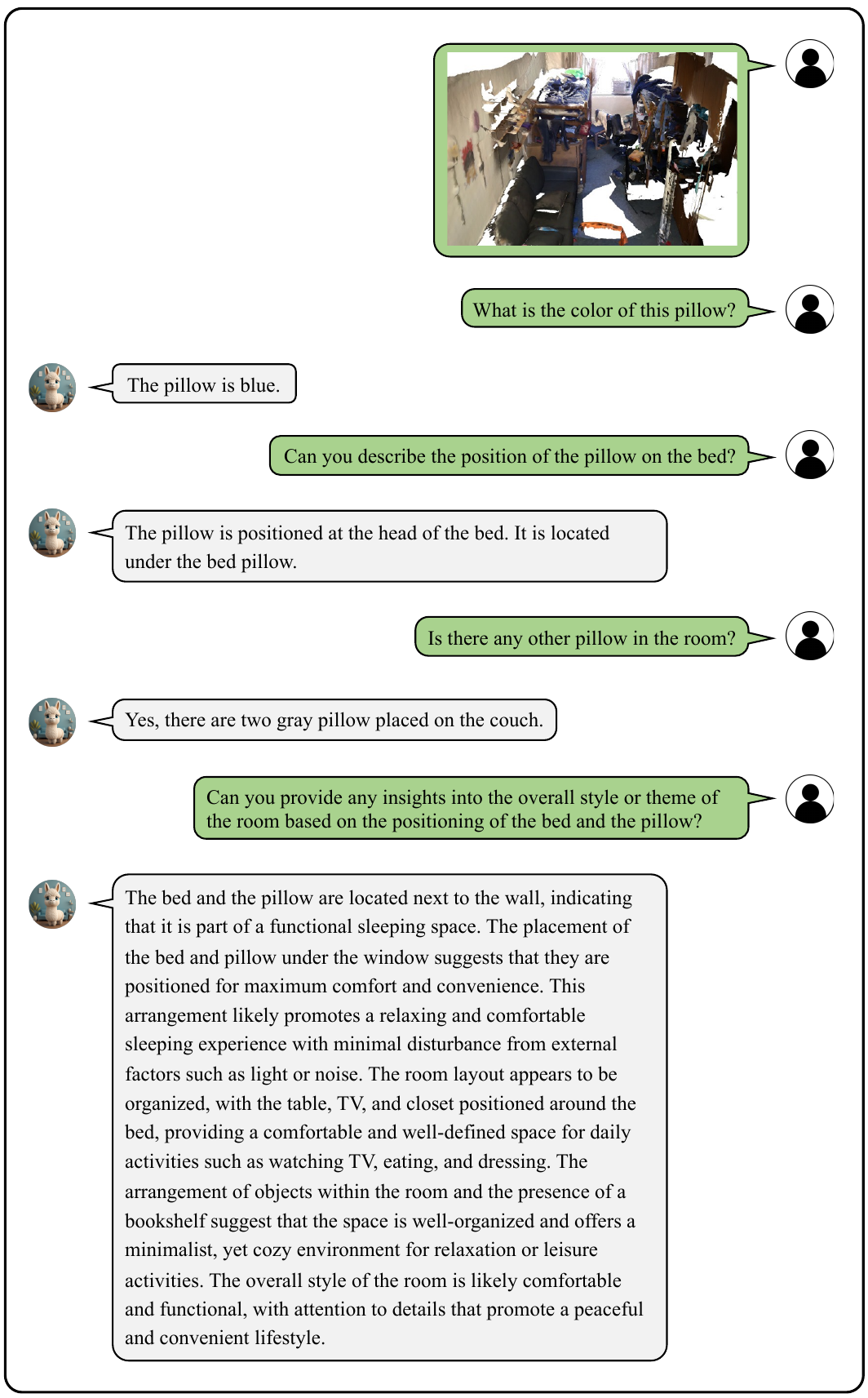}
  \caption{Example 5 of Chat-3D conversation.}
  \label{fig:chat_example5}
\end{figure}

\begin{figure}[tb]
  \centering
  \includegraphics[width=0.85\textwidth]{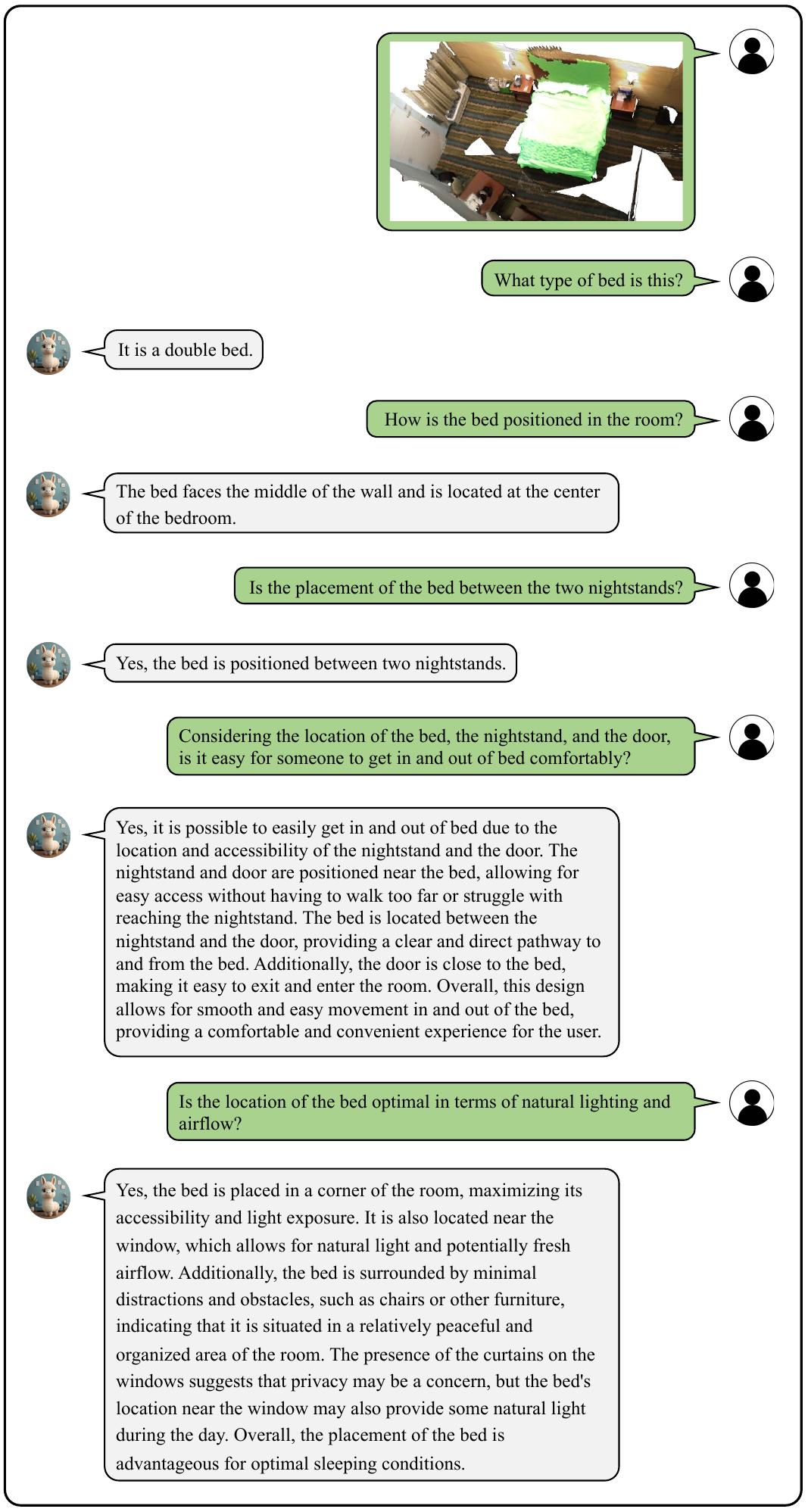}
  \caption{Example 6 of Chat-3D conversation.}
  \label{fig:chat_example6}
\end{figure}

\paragraph{Perception, Reasoning and Dialogue}
We provide several examples of conversations with Chat-3D in \cref{fig:chat_example1,fig:chat_example2,fig:chat_example3,fig:chat_example4,fig:chat_example5,fig:chat_example6}, covering various commonly-seen object types (\textit{e.g.}, table, chair, and bed). In \cref{fig:chat_example1}, Chat-3D shows strong perception capabilities by accurately counting objects, recognizing shapes, and precisely localizing them within the 3D space. In \cref{fig:chat_example3}, Chat-3D demonstrates impressive reasoning capabilities by deducing the cabinet's purpose and evaluating its practicality based on its placement and spatial relationships with surrounding objects. Guided by the object-centric prompt outlined in Section~\ref{sec:arch}, Chat-3D adeptly directs its attention to the specific target object indicated by the user. This enables Chat-3D to maintain focus on the intended subject without being diverted by other similar objects present in the scene. Moreover, the conversational exchanges consistently demonstrate the high-quality dialogue delivered by Chat-3D.

\begin{figure}[tb]
  \centering
  \includegraphics[width=\textwidth]{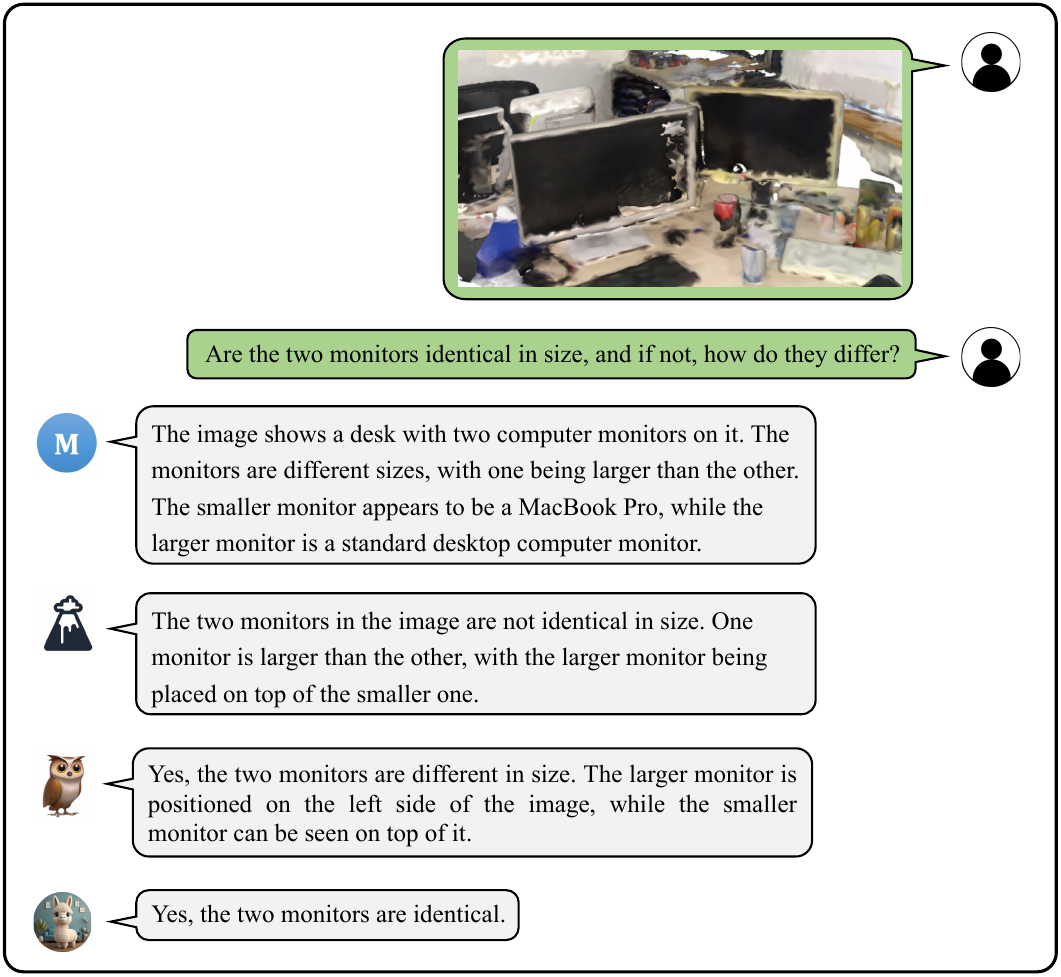}
  \caption{Example 1 of comparison between Chat-3D and 2D Multi-modal LLMs.}
  \label{fig:chat_compare1}
\end{figure}

\begin{figure}[tb]
  \centering
  \includegraphics[width=\textwidth]{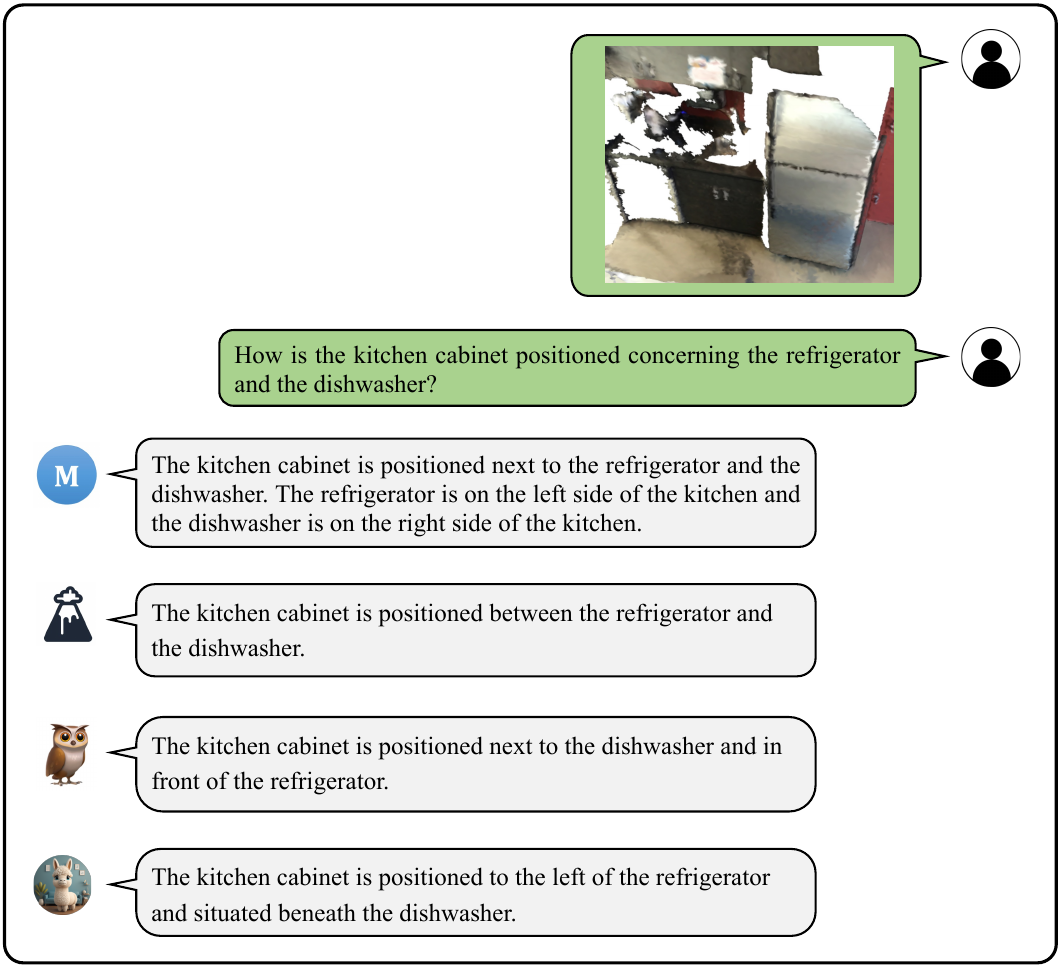}
  \caption{Example 2 of comparison between Chat-3D and 2D Multi-modal LLMs.}
  \label{fig:chat_compare2}
\end{figure}

\begin{figure}[tb]
  \centering
  \includegraphics[width=\textwidth]{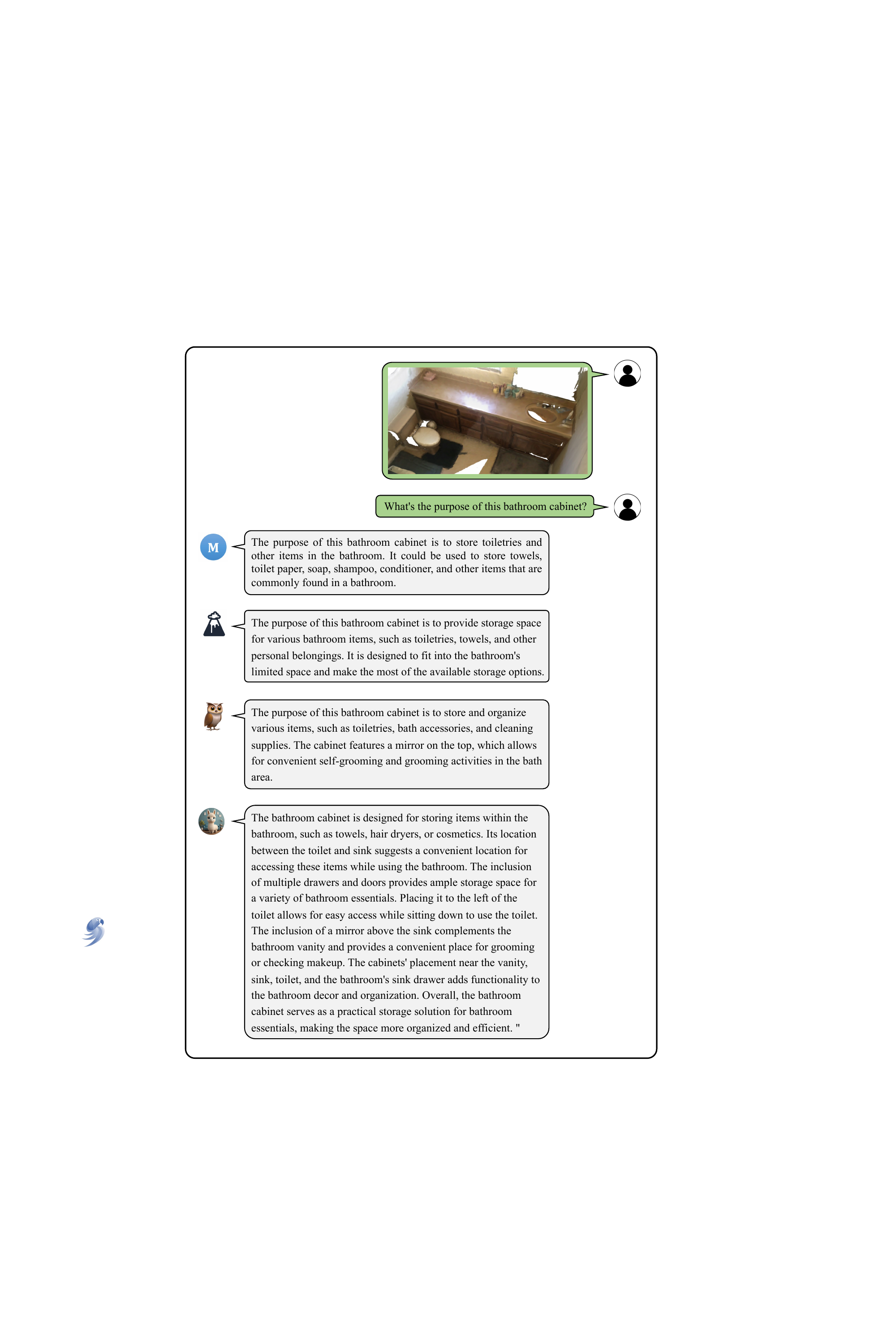}
  \caption{Example 3 of comparison between Chat-3D and 2D Multi-modal LLMs.}
  \label{fig:chat_compare3}
\end{figure}

\begin{figure}[tb]
  \centering
  \includegraphics[width=0.9\textwidth]{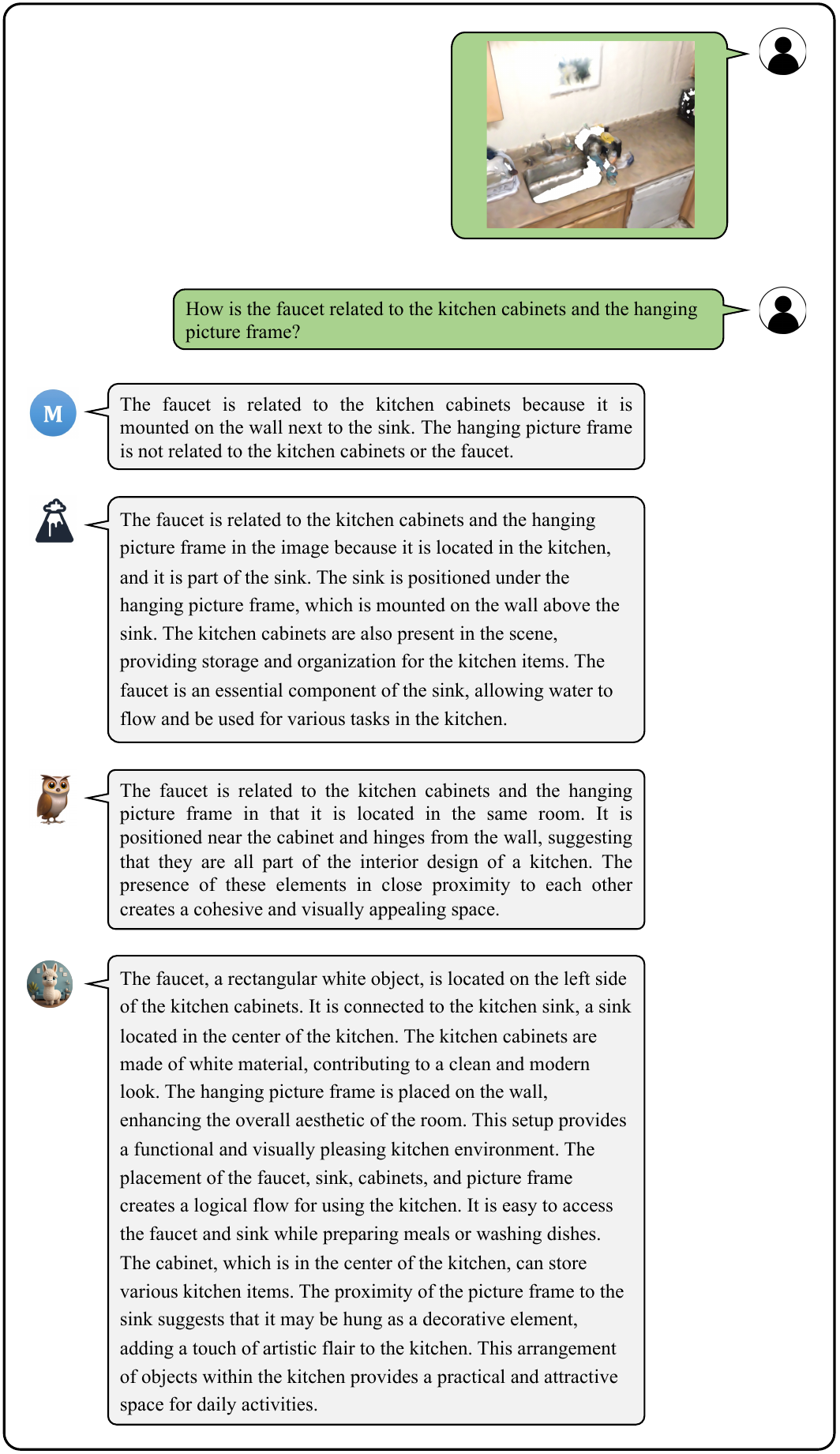}
  \caption{Example 4 of comparison between Chat-3D and 2D Multi-modal LLMs.}
  \label{fig:chat_compare4}
\end{figure}

\paragraph{Comparisons with 2D Multi-modal LLMs}
We compare Chat-3D with MiniGPT-4\raisebox{-0.2ex}{\includegraphics[width=1em]{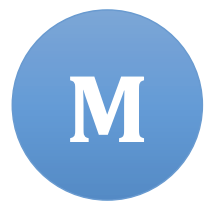}}\cite{zhu2023minigpt}, LLaVA\raisebox{-0.2ex}{\includegraphics[width=1em]{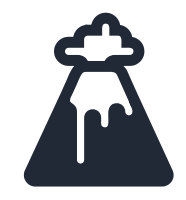}}\cite{liu2023visual}, and mPLUG-owl\raisebox{-0.4ex}{\includegraphics[width=1em]{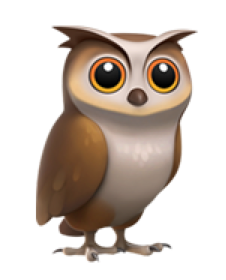}}\cite{ye2023mplug} in \cref{fig:chat_compare1,fig:chat_compare2,fig:chat_compare3,fig:chat_compare4}.
Example 1, depicted in \cref{fig:chat_compare1}, evaluates the model's spatial perception ability in discerning whether both monitors are of identical size. Chat-3D demonstrates accurate identification, while the other 2D models provide incorrect answers due to their limitations in comprehending depth and perspective relationships within the 2D image.
In example 2, presented in \cref{fig:chat_compare2}, the limitations of 2D models are further exposed in their inability to accurately identify the spatial relationships between the target object and its surrounding objects.
Furthermore, the outstanding reasoning prowess of Chat-3D is exemplified through example 3 in \cref{fig:chat_compare3}, showcasing its capacity to deliver a clear and meticulous analysis of the given question. In comparison to 2D models, Chat-3D's analytical prowess shines brightly due to its remarkable aptitude for perceiving and comprehending concepts within the 3D space.

\section{Conclusion}
In this paper, we build the first universal dialogue system for 3D scenes, leveraging the advanced visual perception capabilities of 3D pre-trained models, in conjunction with the powerful reasoning and open-domain conversational abilities of LLMs. To overcome the challenge of limited 3D data availability, we introduce a three-stage training scheme for multi-modal LLMs to progressively transition from learning individual object attributes to capturing complex spatial object relations. Furthermore, we construct a high-quality object-centric 3D instruction dataset and propose a corresponding object-centric prompt approach to facilitate a user-friendly interaction method. Experimental results demonstrate that Chat-3D showcases remarkable capabilities in universal dialogue, spatial reasoning, and the enhancement of external knowledge based on 3D scenes.

\newpage

\bibliographystyle{unsrt}
\bibliography{main}

\begin{thebibliography}{10}

\bibitem{azuma2022scanqa}
Daichi Azuma, Taiki Miyanishi, Shuhei Kurita, and Motoaki Kawanabe.
\newblock Scanqa: 3d question answering for spatial scene understanding.
\newblock In {\em proceedings of the IEEE/CVF conference on computer vision and
  pattern recognition}, pages 19129--19139, 2022.

\bibitem{ma2022sqa3d}
Xiaojian Ma, Silong Yong, Zilong Zheng, Qing Li, Yitao Liang, Song-Chun Zhu,
  and Siyuan Huang.
\newblock Sqa3d: Situated question answering in 3d scenes.
\newblock {\em arXiv preprint arXiv:2210.07474}, 2022.

\bibitem{chen2020scanrefer}
Dave~Zhenyu Chen, Angel~X Chang, and Matthias Nie{\ss}ner.
\newblock Scanrefer: 3d object localization in rgb-d scans using natural
  language.
\newblock In {\em European conference on computer vision}, pages 202--221.
  Springer, 2020.

\bibitem{achlioptas2020referit3d}
Panos Achlioptas, Ahmed Abdelreheem, Fei Xia, Mohamed Elhoseiny, and Leonidas
  Guibas.
\newblock Referit3d: Neural listeners for fine-grained 3d object identification
  in real-world scenes.
\newblock In {\em Computer Vision--ECCV 2020: 16th European Conference,
  Glasgow, UK, August 23--28, 2020, Proceedings, Part I 16}, pages 422--440.
  Springer, 2020.

\bibitem{chen2021scan2cap}
Zhenyu Chen, Ali Gholami, Matthias Nie{\ss}ner, and Angel~X Chang.
\newblock Scan2cap: Context-aware dense captioning in rgb-d scans.
\newblock In {\em Proceedings of the IEEE/CVF conference on computer vision and
  pattern recognition}, pages 3193--3203, 2021.

\bibitem{wang20233drp}
Zehan Wang, Haifeng Huang, Yang Zhao, Linjun Li, Xize Cheng, Yichen Zhu,
  Aoxiong Yin, and Zhou Zhao.
\newblock 3drp-net: 3d relative position-aware network for 3d visual grounding.
\newblock {\em arXiv preprint arXiv:2307.13363}, 2023.

\bibitem{wang2023distilling}
Zehan Wang, Haifeng Huang, Yang Zhao, Linjun Li, Xize Cheng, Yichen Zhu,
  Aoxiong Yin, and Zhou Zhao.
\newblock Distilling coarse-to-fine semantic matching knowledge for weakly
  supervised 3d visual grounding.
\newblock {\em arXiv preprint arXiv:2307.09267}, 2023.

\bibitem{yang2021sat}
Zhengyuan Yang, Songyang Zhang, Liwei Wang, and Jiebo Luo.
\newblock Sat: 2d semantics assisted training for 3d visual grounding.
\newblock In {\em Proceedings of the IEEE/CVF International Conference on
  Computer Vision}, pages 1856--1866, 2021.

\bibitem{jiao2022more}
Yang Jiao, Shaoxiang Chen, Zequn Jie, Jingjing Chen, Lin Ma, and Yu-Gang Jiang.
\newblock More: Multi-order relation mining for dense captioning in 3d scenes.
\newblock In {\em European Conference on Computer Vision}, pages 528--545.
  Springer, 2022.

\bibitem{yuan2022x}
Zhihao Yuan, Xu~Yan, Yinghong Liao, Yao Guo, Guanbin Li, Shuguang Cui, and Zhen
  Li.
\newblock X-trans2cap: Cross-modal knowledge transfer using transformer for 3d
  dense captioning.
\newblock In {\em Proceedings of the IEEE/CVF Conference on Computer Vision and
  Pattern Recognition}, pages 8563--8573, 2022.

\bibitem{parelli2023clip}
Maria Parelli, Alexandros Delitzas, Nikolas Hars, Georgios Vlassis, Sotirios
  Anagnostidis, Gregor Bachmann, and Thomas Hofmann.
\newblock Clip-guided vision-language pre-training for question answering in 3d
  scenes.
\newblock In {\em Proceedings of the IEEE/CVF Conference on Computer Vision and
  Pattern Recognition}, pages 5606--5611, 2023.

\bibitem{yu2022point}
Xumin Yu, Lulu Tang, Yongming Rao, Tiejun Huang, Jie Zhou, and Jiwen Lu.
\newblock Point-bert: Pre-training 3d point cloud transformers with masked
  point modeling.
\newblock In {\em Proceedings of the IEEE/CVF Conference on Computer Vision and
  Pattern Recognition}, pages 19313--19322, 2022.

\bibitem{pang2022masked}
Yatian Pang, Wenxiao Wang, Francis~EH Tay, Wei Liu, Yonghong Tian, and Li~Yuan.
\newblock Masked autoencoders for point cloud self-supervised learning.
\newblock In {\em Computer Vision--ECCV 2022: 17th European Conference, Tel
  Aviv, Israel, October 23--27, 2022, Proceedings, Part II}, pages 604--621.
  Springer, 2022.

\bibitem{wang2021unsupervised}
Hanchen Wang, Qi~Liu, Xiangyu Yue, Joan Lasenby, and Matt~J Kusner.
\newblock Unsupervised point cloud pre-training via occlusion completion.
\newblock In {\em Proceedings of the IEEE/CVF international conference on
  computer vision}, pages 9782--9792, 2021.

\bibitem{zhang2022point}
Renrui Zhang, Ziyu Guo, Peng Gao, Rongyao Fang, Bin Zhao, Dong Wang, Yu~Qiao,
  and Hongsheng Li.
\newblock Point-m2ae: multi-scale masked autoencoders for hierarchical point
  cloud pre-training.
\newblock {\em Advances in neural information processing systems},
  35:27061--27074, 2022.

\bibitem{xue2023ulip}
Le~Xue, Mingfei Gao, Chen Xing, Roberto Mart{\'\i}n-Mart{\'\i}n, Jiajun Wu,
  Caiming Xiong, Ran Xu, Juan~Carlos Niebles, and Silvio Savarese.
\newblock Ulip: Learning a unified representation of language, images, and
  point clouds for 3d understanding.
\newblock In {\em Proceedings of the IEEE/CVF Conference on Computer Vision and
  Pattern Recognition}, pages 1179--1189, 2023.

\bibitem{liu2023openshape}
Minghua Liu, Ruoxi Shi, Kaiming Kuang, Yinhao Zhu, Xuanlin Li, Shizhong Han,
  Hong Cai, Fatih Porikli, and Hao Su.
\newblock Openshape: Scaling up 3d shape representation towards open-world
  understanding.
\newblock {\em arXiv preprint arXiv:2305.10764}, 2023.

\bibitem{chiang2023vicuna}
Wei-Lin Chiang, Zhuohan Li, Zi~Lin, Ying Sheng, Zhanghao Wu, Hao Zhang, Lianmin
  Zheng, Siyuan Zhuang, Yonghao Zhuang, Joseph~E Gonzalez, et~al.
\newblock Vicuna: An open-source chatbot impressing gpt-4 with 90\%* chatgpt
  quality.
\newblock {\em See https://vicuna. lmsys. org (accessed 14 April 2023)}, 2023.

\bibitem{openai2023gpt4}
OpenAI.
\newblock Gpt-4 technical report, 2023.

\bibitem{touvron2023llama}
Hugo Touvron, Thibaut Lavril, Gautier Izacard, Xavier Martinet, Marie-Anne
  Lachaux, Timoth{\'e}e Lacroix, Baptiste Rozi{\`e}re, Naman Goyal, Eric
  Hambro, Faisal Azhar, et~al.
\newblock Llama: Open and efficient foundation language models.
\newblock {\em arXiv preprint arXiv:2302.13971}, 2023.

\bibitem{chowdhery2022palm}
Aakanksha Chowdhery, Sharan Narang, Jacob Devlin, Maarten Bosma, Gaurav Mishra,
  Adam Roberts, Paul Barham, Hyung~Won Chung, Charles Sutton, Sebastian
  Gehrmann, et~al.
\newblock Palm: Scaling language modeling with pathways.
\newblock {\em arXiv preprint arXiv:2204.02311}, 2022.

\bibitem{li2023videochat}
KunChang Li, Yinan He, Yi~Wang, Yizhuo Li, Wenhai Wang, Ping Luo, Yali Wang,
  Limin Wang, and Yu~Qiao.
\newblock Videochat: Chat-centric video understanding.
\newblock {\em arXiv preprint arXiv:2305.06355}, 2023.

\bibitem{liu2023visual}
Haotian Liu, Chunyuan Li, Qingyang Wu, and Yong~Jae Lee.
\newblock Visual instruction tuning.
\newblock {\em arXiv preprint arXiv:2304.08485}, 2023.

\bibitem{zhao2023bubogpt}
Yang Zhao, Zhijie Lin, Daquan Zhou, Zilong Huang, Jiashi Feng, and Bingyi Kang.
\newblock Bubogpt: Enabling visual grounding in multi-modal llms.
\newblock {\em arXiv preprint arXiv:2307.08581}, 2023.

\bibitem{zhang2023video}
Hang Zhang, Xin Li, and Lidong Bing.
\newblock Video-llama: An instruction-tuned audio-visual language model for
  video understanding.
\newblock {\em arXiv preprint arXiv:2306.02858}, 2023.

\bibitem{zhu2023minigpt}
Deyao Zhu, Jun Chen, Xiaoqian Shen, Xiang Li, and Mohamed Elhoseiny.
\newblock Minigpt-4: Enhancing vision-language understanding with advanced
  large language models.
\newblock {\em arXiv preprint arXiv:2304.10592}, 2023.

\bibitem{lin2014microsoft}
Tsung-Yi Lin, Michael Maire, Serge Belongie, James Hays, Pietro Perona, Deva
  Ramanan, Piotr Doll{\'a}r, and C~Lawrence Zitnick.
\newblock Microsoft coco: Common objects in context.
\newblock In {\em Computer Vision--ECCV 2014: 13th European Conference, Zurich,
  Switzerland, September 6-12, 2014, Proceedings, Part V 13}, pages 740--755.
  Springer, 2014.

\bibitem{sharma2018conceptual}
Piyush Sharma, Nan Ding, Sebastian Goodman, and Radu Soricut.
\newblock Conceptual captions: A cleaned, hypernymed, image alt-text dataset
  for automatic image captioning.
\newblock In {\em Proceedings of the 56th Annual Meeting of the Association for
  Computational Linguistics (Volume 1: Long Papers)}, pages 2556--2565, 2018.

\bibitem{changpinyo2021conceptual}
Soravit Changpinyo, Piyush Sharma, Nan Ding, and Radu Soricut.
\newblock Conceptual 12m: Pushing web-scale image-text pre-training to
  recognize long-tail visual concepts.
\newblock In {\em Proceedings of the IEEE/CVF Conference on Computer Vision and
  Pattern Recognition}, pages 3558--3568, 2021.

\bibitem{schuhmann2021laion}
Christoph Schuhmann, Richard Vencu, Romain Beaumont, Robert Kaczmarczyk,
  Clayton Mullis, Aarush Katta, Theo Coombes, Jenia Jitsev, and Aran
  Komatsuzaki.
\newblock Laion-400m: Open dataset of clip-filtered 400 million image-text
  pairs.
\newblock {\em arXiv preprint arXiv:2111.02114}, 2021.

\bibitem{schuhmann2022laion}
Christoph Schuhmann, Romain Beaumont, Richard Vencu, Cade Gordon, Ross
  Wightman, Mehdi Cherti, Theo Coombes, Aarush Katta, Clayton Mullis, Mitchell
  Wortsman, et~al.
\newblock Laion-5b: An open large-scale dataset for training next generation
  image-text models.
\newblock {\em Advances in Neural Information Processing Systems},
  35:25278--25294, 2022.

\bibitem{bain2021frozen}
Max Bain, Arsha Nagrani, G{\"u}l Varol, and Andrew Zisserman.
\newblock Frozen in time: A joint video and image encoder for end-to-end
  retrieval.
\newblock In {\em Proceedings of the IEEE/CVF International Conference on
  Computer Vision}, pages 1728--1738, 2021.

\bibitem{miech2019howto100m}
Antoine Miech, Dimitri Zhukov, Jean-Baptiste Alayrac, Makarand Tapaswi, Ivan
  Laptev, and Josef Sivic.
\newblock Howto100m: Learning a text-video embedding by watching hundred
  million narrated video clips.
\newblock In {\em Proceedings of the IEEE/CVF international conference on
  computer vision}, pages 2630--2640, 2019.

\bibitem{xu2016msr}
Jun Xu, Tao Mei, Ting Yao, and Yong Rui.
\newblock Msr-vtt: A large video description dataset for bridging video and
  language.
\newblock In {\em Proceedings of the IEEE conference on computer vision and
  pattern recognition}, pages 5288--5296, 2016.

\bibitem{radford2021learning}
Alec Radford, Jong~Wook Kim, Chris Hallacy, Aditya Ramesh, Gabriel Goh,
  Sandhini Agarwal, Girish Sastry, Amanda Askell, Pamela Mishkin, Jack Clark,
  et~al.
\newblock Learning transferable visual models from natural language
  supervision.
\newblock In {\em International conference on machine learning}, pages
  8748--8763. PMLR, 2021.

\bibitem{cherti2023reproducible}
Mehdi Cherti, Romain Beaumont, Ross Wightman, Mitchell Wortsman, Gabriel
  Ilharco, Cade Gordon, Christoph Schuhmann, Ludwig Schmidt, and Jenia Jitsev.
\newblock Reproducible scaling laws for contrastive language-image learning.
\newblock In {\em Proceedings of the IEEE/CVF Conference on Computer Vision and
  Pattern Recognition}, pages 2818--2829, 2023.

\bibitem{zhang2023learning}
Renrui Zhang, Liuhui Wang, Yu~Qiao, Peng Gao, and Hongsheng Li.
\newblock Learning 3d representations from 2d pre-trained models via
  image-to-point masked autoencoders.
\newblock In {\em Proceedings of the IEEE/CVF Conference on Computer Vision and
  Pattern Recognition}, pages 21769--21780, 2023.

\bibitem{dong2023act}
Runpei Dong, Zekun Qi, Linfeng Zhang, Junbo Zhang, Jianjian Sun, Zheng Ge,
  Li~Yi, and Kaisheng Ma.
\newblock Autoencoders as cross-modal teachers: Can pretrained 2d image
  transformers help 3d representation learning?
\newblock In {\em The Eleventh International Conference on Learning
  Representations (ICLR)}, 2023.

\bibitem{dosovitskiy2020image}
Alexey Dosovitskiy, Lucas Beyer, Alexander Kolesnikov, Dirk Weissenborn,
  Xiaohua Zhai, Thomas Unterthiner, Mostafa Dehghani, Matthias Minderer, Georg
  Heigold, Sylvain Gelly, et~al.
\newblock An image is worth 16x16 words: Transformers for image recognition at
  scale.
\newblock {\em arXiv preprint arXiv:2010.11929}, 2020.

\bibitem{he2016deep}
Kaiming He, Xiangyu Zhang, Shaoqing Ren, and Jian Sun.
\newblock Deep residual learning for image recognition.
\newblock In {\em Proceedings of the IEEE conference on computer vision and
  pattern recognition}, pages 770--778, 2016.

\bibitem{kazemzadeh2014referitgame}
Sahar Kazemzadeh, Vicente Ordonez, Mark Matten, and Tamara Berg.
\newblock Referitgame: Referring to objects in photographs of natural scenes.
\newblock In {\em Proceedings of the 2014 conference on empirical methods in
  natural language processing (EMNLP)}, pages 787--798, 2014.

\bibitem{krishna2017dense}
Ranjay Krishna, Kenji Hata, Frederic Ren, Li~Fei-Fei, and Juan Carlos~Niebles.
\newblock Dense-captioning events in videos.
\newblock In {\em Proceedings of the IEEE international conference on computer
  vision}, pages 706--715, 2017.

\bibitem{goyal2017making}
Yash Goyal, Tejas Khot, Douglas Summers-Stay, Dhruv Batra, and Devi Parikh.
\newblock Making the v in vqa matter: Elevating the role of image understanding
  in visual question answering.
\newblock In {\em Proceedings of the IEEE conference on computer vision and
  pattern recognition}, pages 6904--6913, 2017.

\bibitem{antol2015vqa}
Stanislaw Antol, Aishwarya Agrawal, Jiasen Lu, Margaret Mitchell, Dhruv Batra,
  C~Lawrence Zitnick, and Devi Parikh.
\newblock Vqa: Visual question answering.
\newblock In {\em Proceedings of the IEEE international conference on computer
  vision}, pages 2425--2433, 2015.

\bibitem{grauman2022ego4d}
Kristen Grauman, Andrew Westbury, Eugene Byrne, Zachary Chavis, Antonino
  Furnari, Rohit Girdhar, Jackson Hamburger, Hao Jiang, Miao Liu, Xingyu Liu,
  et~al.
\newblock Ego4d: Around the world in 3,000 hours of egocentric video.
\newblock In {\em Proceedings of the IEEE/CVF Conference on Computer Vision and
  Pattern Recognition}, pages 18995--19012, 2022.

\bibitem{li2022blip}
Junnan Li, Dongxu Li, Caiming Xiong, and Steven Hoi.
\newblock Blip: Bootstrapping language-image pre-training for unified
  vision-language understanding and generation.
\newblock In {\em International Conference on Machine Learning}, pages
  12888--12900. PMLR, 2022.

\bibitem{li2023blip}
Junnan Li, Dongxu Li, Silvio Savarese, and Steven Hoi.
\newblock Blip-2: Bootstrapping language-image pre-training with frozen image
  encoders and large language models.
\newblock {\em arXiv preprint arXiv:2301.12597}, 2023.

\bibitem{li2021align}
Junnan Li, Ramprasaath Selvaraju, Akhilesh Gotmare, Shafiq Joty, Caiming Xiong,
  and Steven Chu~Hong Hoi.
\newblock Align before fuse: Vision and language representation learning with
  momentum distillation.
\newblock {\em Advances in neural information processing systems},
  34:9694--9705, 2021.

\bibitem{lin2022swinbert}
Kevin Lin, Linjie Li, Chung-Ching Lin, Faisal Ahmed, Zhe Gan, Zicheng Liu,
  Yumao Lu, and Lijuan Wang.
\newblock Swinbert: End-to-end transformers with sparse attention for video
  captioning.
\newblock In {\em Proceedings of the IEEE/CVF Conference on Computer Vision and
  Pattern Recognition}, pages 17949--17958, 2022.

\bibitem{deng2021transvg}
Jiajun Deng, Zhengyuan Yang, Tianlang Chen, Wengang Zhou, and Houqiang Li.
\newblock Transvg: End-to-end visual grounding with transformers.
\newblock In {\em Proceedings of the IEEE/CVF International Conference on
  Computer Vision}, pages 1769--1779, 2021.

\bibitem{wang2023scene}
Zehan Wang, Yang Zhao, Haifeng Huang, Yan Xia, and Zhou Zhao.
\newblock Scene-robust natural language video localization via learning
  domain-invariant representations.
\newblock In {\em Findings of the Association for Computational Linguistics:
  ACL 2023}, pages 144--160, 2023.

\bibitem{jiang2020pointgroup}
Li~Jiang, Hengshuang Zhao, Shaoshuai Shi, Shu Liu, Chi-Wing Fu, and Jiaya Jia.
\newblock Pointgroup: Dual-set point grouping for 3d instance segmentation.
\newblock In {\em Proceedings of the IEEE/CVF conference on computer vision and
  Pattern recognition}, pages 4867--4876, 2020.

\bibitem{misra2021end}
Ishan Misra, Rohit Girdhar, and Armand Joulin.
\newblock An end-to-end transformer model for 3d object detection.
\newblock In {\em Proceedings of the IEEE/CVF International Conference on
  Computer Vision}, pages 2906--2917, 2021.

\bibitem{qi2019deep}
Charles~R Qi, Or~Litany, Kaiming He, and Leonidas~J Guibas.
\newblock Deep hough voting for 3d object detection in point clouds.
\newblock In {\em proceedings of the IEEE/CVF International Conference on
  Computer Vision}, pages 9277--9286, 2019.

\bibitem{dai2017scannet}
Angela Dai, Angel~X Chang, Manolis Savva, Maciej Halber, Thomas Funkhouser, and
  Matthias Nie{\ss}ner.
\newblock Scannet: Richly-annotated 3d reconstructions of indoor scenes.
\newblock In {\em Proceedings of the IEEE conference on computer vision and
  pattern recognition}, pages 5828--5839, 2017.

\bibitem{chang2015shapenet}
Angel~X Chang, Thomas Funkhouser, Leonidas Guibas, Pat Hanrahan, Qixing Huang,
  Zimo Li, Silvio Savarese, Manolis Savva, Shuran Song, Hao Su, et~al.
\newblock Shapenet: An information-rich 3d model repository.
\newblock {\em arXiv preprint arXiv:1512.03012}, 2015.

\bibitem{uy2019revisiting}
Mikaela~Angelina Uy, Quang-Hieu Pham, Binh-Son Hua, Thanh Nguyen, and Sai-Kit
  Yeung.
\newblock Revisiting point cloud classification: A new benchmark dataset and
  classification model on real-world data.
\newblock In {\em Proceedings of the IEEE/CVF international conference on
  computer vision}, pages 1588--1597, 2019.

\bibitem{deitke2023objaverse}
Matt Deitke, Dustin Schwenk, Jordi Salvador, Luca Weihs, Oscar Michel, Eli
  VanderBilt, Ludwig Schmidt, Kiana Ehsani, Aniruddha Kembhavi, and Ali
  Farhadi.
\newblock Objaverse: A universe of annotated 3d objects.
\newblock In {\em Proceedings of the IEEE/CVF Conference on Computer Vision and
  Pattern Recognition}, pages 13142--13153, 2023.

\bibitem{pointbind}
Ziyu Guo.
\newblock Point-bind: Align 3d point clouds with multi-modalities.
\newblock \url{https://github.com/ZrrSkywalker/Point-Bind}.

\bibitem{ye2023mplug}
Qinghao Ye, Haiyang Xu, Guohai Xu, Jiabo Ye, Ming Yan, Yiyang Zhou, Junyang
  Wang, Anwen Hu, Pengcheng Shi, Yaya Shi, et~al.
\newblock mplug-owl: Modularization empowers large language models with
  multimodality.
\newblock {\em arXiv preprint arXiv:2304.14178}, 2023.

\end{thebibliography}

\end{document}